# Spine-Inspired Continuum Soft Exoskeleton for Stoop Lifting Assistance

Xiaolong Yang, Tzu-Hao Huang, Hang Hu, Shuangyue Yu, Sainan Zhang, Xianlian Zhou,
Alessandra Carriero, Guang Yue, and Hao Su, *Member, IEEE*

*Abstract*—Back injuries are the most prevalent work-related musculoskeletal disorders and represent a major cause of disability. Although innovations in wearable robots aim to alleviate this hazard, the majority of existing exoskeletons are obtrusive because the rigid linkage design limits natural movement, thus causing ergonomic risk. Moreover, these existing systems are typically only suitable for one type of movement assistance, not ubiquitous for a wide variety of activities. To fill in this gap, this paper presents a new wearable robot design approach continuum soft exoskeleton. This spine-inspired wearable robot is unobtrusive and assists both squat and stoops while not impeding walking motion. To tackle the challenge of the unique anatomy of spine that is inappropriate to be simplified as a single degree of freedom joint, our robot is conformal to human anatomy and it can reduce multiple types of forces along the human spine such as the spinae muscle force, shear, and compression force of the lumbar vertebrae. We derived kinematics and kinetics models of this mechanism and established an analytical biomechanics model of human-robot interaction. Quantitative analysis of disc compression force, disc shear force and muscle force was performed in simulation. We further developed a virtual impedance control strategy to deliver force control and compensate hysteresis of Bowden cable transmission. The feasibility of the prototype was experimentally tested on three healthy subjects. The root mean square error of force tracking is 6.63 N (3.3 % of the 200N peak force) and it demonstrated that it can actively control the stiffness to the desired value. This continuum soft exoskeleton represents a feasible solution with the potential to reduce back pain for multiple activities and multiple forces along the human spine.

## I. INTRODUCTION

Workplace-related injuries are estimated to cost $250 billion every year in the U.S., with 89 million workers exposed to the risk of preventable injuries. Back injuries, which represented 17.3% of all injuries in 2016, are the most prevalent work-related musculoskeletal disorders [1]. Wearable robots present an attractive solution to mitigate ergonomic risk factors and reduce musculoskeletal loading for workers who perform lifting. Over the last two decades, various studies have demonstrated that industrial exoskeletons can decrease total work, fatigue, and load while increasing productivity and work quality [2-3]. Another prominent field where exoskeletons have been heralded as a promising technology is medical rehabilitation focusing on walking assistance [4-12]. Recently, the feasibility of exoskeleton-assisted energetics reduction has been demonstrated in walkers [11-13], post-stroke individuals with paretic limbs [14], load carriers [15], children with cerebral palsy [16], and joggers [17].

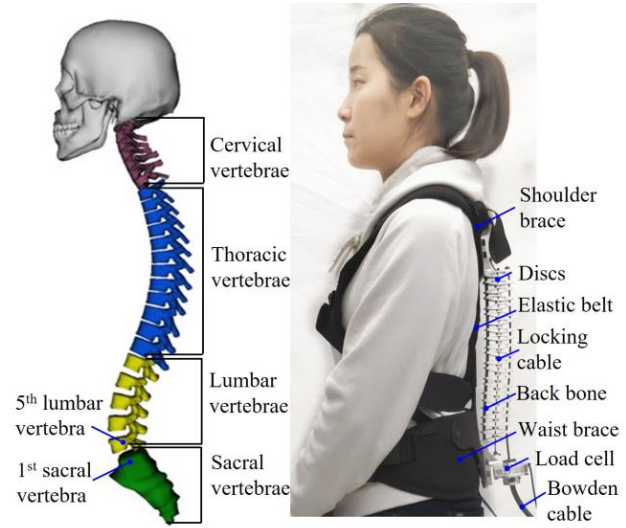

Fig. 1 The spine-inspired back exoskeleton is composed of a continuum mechanism, wearable structure (shoulder and waist braces) and a tethered actuation platform. Since the spine-inspired soft exoskeleton is a hyper-redundant continuum mechanism that continuously bends, this under-actuation robot provides assistive force while being conformal to the anatomy of the human spine. It has no limit on human natural motion, allowing the wearer to flex, lateral flex and rotate.

The key challenges of back-support exoskeletons lie in the stringent requirements [18] that need to augment human capability in different postures (squat and stoop lifting), during different activities (e.g. walking and lifting), for multiple joints (e.g. erector spinae muscle, and lumbar vertebral compression and shear forces). Rigid exoskeletons rely on transmission mechanisms made of rigid components [19], which typically limit the natural movement of wearers. Toxiri et al. [20] developed a powered back-support exoskeleton that reduced 30% muscular activity at the lumbar spine. Naf et al. [21] proposed a passive back exoskeleton with a 25% increase of the range of motion of the trunk in the sagittal plane compared with the rigid powered design [20].

Soft exoskeletons use soft materials and employ pneumatics or cable-driven transmissions to assist limb movement. Pneumatic actuation [25][26] shows promise as it avoids joint misalignment issue that limits human motion.

X. Yang, T. Huang, H. Hu, S. Yu, S. Zhang, and H. Su are with the Lab of Biomechatronics and Intelligent Robotics, Department of Mechanical Engineering, The City University of New York, City College, New York, NY, 10031, US. X. Zhou is with Department of Biomedical Engineering, New Jersey Institute of Technology, New Jersey, NJ 07102, US. A. Carriero is with the Department of Biomedical Engineering, The City University of New York, City College, 10031, US. G. Yue is with Kessler Foundation, New Jersey, NJ 07936, US. Y. †Corresponding author. Email: hao.su@ccny.cuny.edu

However, it relies on a tethered air compressor. Thus it is challenging to develop portable exoskeletons with pneumatic actuation. Textile soft exosuit [24] represents a cable-driven mechanism based electric actuation. This innovative solution has demonstrated benefits for ankle [25] and hip [26] augmentation. However, there is no work to study cable-driven soft exoskeleton for the back joint assistance. Moreover, the unique anatomy of human back presents new challenges for wearable robot design, as the human spine is composed of 23 intervertebral discs that cannot be approximated as one degree of freedom (DOF) mechanism like the lower limb joints. This necessitates new solutions for robot design, sensing, and control to achieve all functional requirements.

To address the aforementioned challenges, we present a spine-inspired continuum soft exoskeleton (Fig. 1) that reduces spine loadings while not limiting natural movement. Thanks to its hyper-redundant elastic wearable structure that continuously bends [27], this continuum robot is conformal and unobtrusive to human back anatomy with the potential to overcome the limitations in terms of ergonomics [20] and range of motion [21]. The contribution of this paper is a bio-inspired exoskeleton design and biomechanics modeling of human-robot interaction that reduce spine loadings across multiple vertebral joints (spinae muscle force, shear, and compression force) along the back for multiple lifting activities without limiting natural movement. This continuum soft exoskeleton can assist both squat and stoops lifting. This paper focuses on the design and modeling of the robot for stoop assistance with experimental validation.

## II. DESIGN OF CONTINUUM SOFT EXOSKELETON

There are two types of lifting postures, namely squat and stoop with the latter being more energy economic. Our spine-inspired continuum soft exoskeleton aims to 1) assist both stoop and squat lifting; 2) reduce loadings of multiple joints (e.g. erector spine muscle, and lumbar vertebral compression and shear forces). Currently, state of the art back exoskeleton design [20] is not able to reduce all three spinal loadings. Our new exoskeleton performances will be examined in stoop lifting in this paper.

### A. Design Requirements of Back Exoskeletons for Lifting Assistance

The design requirements of the back exoskeleton consider both kinematics and kinetics of human-robot interaction. The stoop lifting induces extension and flexion of the lumbar joints with 70° in the sagittal plane. The robot should not limit the natural motion of a user when wearing the exoskeleton, i.e. the lateral flexion of 20° in the frontal plane and rotation of 90° in the transverse plane. Biomechanics analysis reveals that 250N of the exoskeleton force perpendicular to the back can decrease 30% of the lumbar compression force at the lumbosacral 5th lumbar and 1st sacral (L5/S1) joint while a 15 kg load is lifted.

### B. Spine-Inspired Continuum Mechanism

For the design of back support exoskeletons, the degrees of freedom of the lumbar spine that is composed of 23 intervertebral discs should be taken into account. Toxiri et al. [20] proposed a rigid revolute joint mechanism to provide flexion and extension assistance in the sagittal plane. It demonstrated the effectiveness of muscular activity reduction, but their structure does not conform to the human back.

To address this limitation, the proposed robot leverages on a hyper-redundant continuum mechanism that continuously bends. As shown in Fig. 1, the back exoskeleton is composed of a continuum mechanism, wearable structure (shoulder and waist braces) and a tethered actuation platform. Each of the twenty segments in the spinal structure of the robot is comprised of a disc that pivots on a ball and socket joint. Each disc is made of two spherical cups putting together. As shown in Fig. 1, the exoskeleton assists human in stoop lifting, while has no limit on human natural motion, allowing the wearer to flex forward and laterally, and rotate. Our continuum mechanism is cable-driven; thus it can only be pulled. This is different from the work in [28] as it is nitinol tube based and allows both pull and push motion. That device was made for a minimally invasive surgery requiring a relatively small force (10 N level), while our wearable robot is able to generate large forces (up to 200 N). A cable is threaded through holes at the edges of the discs and when the actuator motor pulls the cable, the discs rotate about the ball joint acting as levers. The segmented nature of the spinal structure also makes the robot conform to the curvature of a wearer's back. A cable passes through a customized load cell at the bottom of the spinal structure to measure the cable tension. When the cable is pulled, the top disc pulls the human back. Another cable placed in the center of the discs ensures the overall mechanism integration and tightly coupled together. An elastic belt is used to connect the shoulder brace and the waist brace. This back exoskeleton provides the assistive force and permits a large range of motion in the sagittal, frontal and transverse planes.

### C. Cable Transmission

As the actuation forces of the human spine are provided by erector spinae muscles adjacent to the spinal column, our robot uses a cable to actuate the spine-inspired continuum mechanism. Another advantage of the cable-driven method is that during lifting, its actuators are in the proximity of the human center of mass to minimize energetic cost due to device mass. Finally, during lifting, the exoskeleton should pull the human to its erect position. Therefore, we designed one cable to pull the top disc of the continuum mechanism to assist the human during erection in the sagittal plane.

## III. MODELING OF CONTINUUM SOFT EXOSKELETON

This section derives kinematics and kinetics modeling for design optimization, a range of motion analysis, and force analysis.

### A. Kinematics of Continuum Soft Exoskeleton

Kinematics analysis is carried out to characterize the motion of the mechanism and optimize the geometric parameters of the discs to satisfy the kinematic requirements of the back exoskeleton. The configuration of the back exoskeleton is determined by the accumulated rotations of all discs, as shown in Fig. 2 (a). The $i+1^{th}$ disc's pose with respect to the $i^{th}$ disc can be represented by a homogeneous transformation

$$\mathbf{T}_{i+1} = \mathbf{Rot}_x(\varphi_{i+1})\mathbf{Rot}_y(\theta_{i+1})\mathbf{Rot}_z(\psi_{i+1})\mathbf{Tran}(l) \quad (1)$$

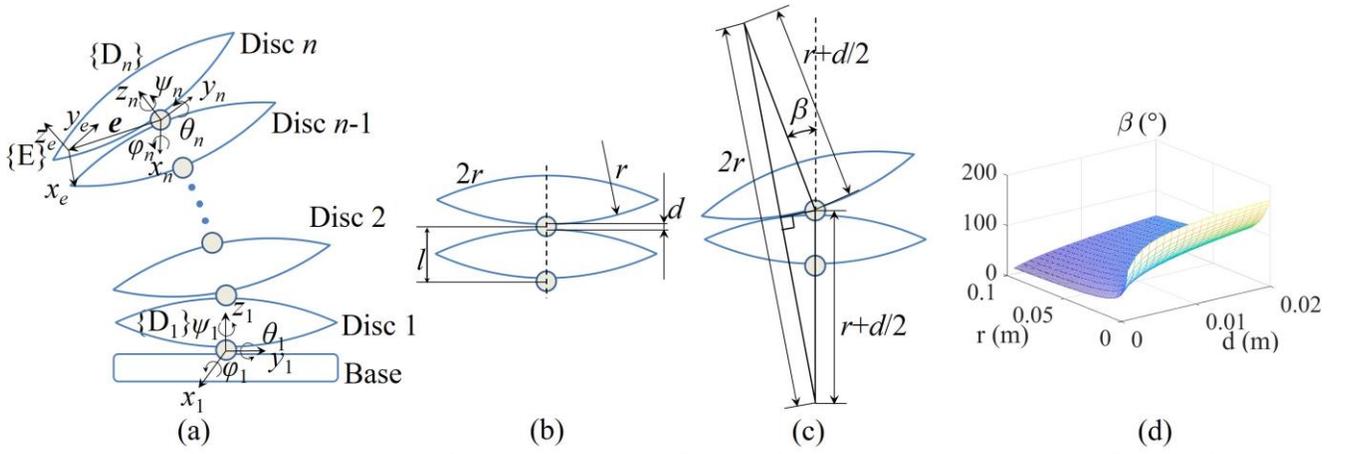

Fig. 2 Kinematics analysis. (a) Accumulated rotations of all the discs. (b) Initial configuration of two adjacent discs. (c) An extreme configuration of two adjacent discs. (d) Variation of the range of the motion β (maximal rotation angle between two neighboring discs) with respect to the geometric parameters r (radius of the disc) and d (distance between the neighboring discs). The range of motion $\beta$ is designed by adjusting the geometric parameters $r$ and $d$.

Coordinate frame {E} is assigned at the connecting point of the end effector (i.e. the distal disc $n$) at the shoulder brace. The pose transformation of the mechanism can be calculated by

$$\mathbf{T}_E = \mathbf{T}_1\mathbf{T}_2\cdots\mathbf{T}_n\mathbf{Tran}(\mathbf{e}) \quad (2)$$

$\varphi_{i+1}, \theta_{i+1}, \psi_{i+1}$ are the rotation angles of disc $i+1$ with respect to disc $i$ in the sagittal, frontal, transverse planes respectively. $\mathbf{l} = (0 \ \ 0 \ \ l)^T$ is the distance vector between two neighboring discs. $Rot_x, Rot_y, Rot_z$ are 4×4 homogeneous transformation matrices representing rotations around $x$, $y$ and $z$-axes, respectively. Tran is a 4×4 homogeneous translation matrix.

To ensure that the exoskeleton conforms to human back anatomy, the range of motion of the exoskeleton should satisfy the requirements specified in Section II. From (2), we see that the overall range of motion is the accumulation of ranges of motion of individual discs. The range of motion of one disc with respect to the adjacent disc depends on the geometric parameters of the disc and the spherical joint in between. When the disc rotates from the initial configuration to the extreme configuration (because of mechanical position limit), as shown in Fig. 2 (b) and (c), the maximal rotation angle, $\beta$, can be calculated by

$$\beta = \pi - 2\arcsin\left(r/(r+d/2)\right) \quad (3)$$

In (3), $r$ denotes the radius of the disc, $d$ denotes the distance between the neighboring discs. $l$ denotes the distance between the centers of two neighboring spherical joints.

We can design the range of motion, $\beta$, by adjusting the parameters $r$ and $d$. As shown in Fig. 2 (d), $\beta$ decreases as $r$ increases, whereas $\beta$ increases as $d$ increases. To keep a low-profile of the exoskeleton, we set $r \in (0, 0.1]$m and $d \in (0, 0.02]$m to observe the variation of $\beta$, as shown in Fig. 2 (d). $\beta$ decreases significantly when $r$ is close to zero and it varies smoothly when $r$ is much larger than $d$. $\beta$ increases slowly as $d$ increase in the whole range. In this paper, we designed $\beta$ as 20°, such that the motion requirement, i.e. flexion of 70° in the sagital plane, the lateral flexion of 20° in the frontal plane and rotation of 90° in the transverse plane, is easy to reach when the amount of discs is more than 6 and the mechanical design is available as well. According to (3), there exist infinite solutions to achieve a certain $\beta$. To make the disc has a low profile and sufficient mechanical strength, we choose $r = 0.07$m and $d = 0.00216$m to obtain $\beta = 20°$.

*B. Kinetics of Continuum Soft Exoskeleton*

To understand the force relation of the under-actuated continuum mechanism, a kinetic model is developed to present the rationale of the design. The force characteristics and advantages of force transmission are discussed as well. The proposed exoskeleton is comprised of serially connected disks with tendon passing through. Each pair of neighboring disks form a three-DOF spherical joint while all disks are constrained by an elastic backbone to keep balance. It is a hyper-redundant mechanism with compliance, and one could represent its configuration using 3$n$ degrees of freedom for $n$ disks [29] or using curvature profile functions to approximate [31, 34]. The mechanism is underactuated, and its configuration (shape) is determined by the one actuator input and the external loads from environments (the human subject). This degree of freedom redundancy is useful to accommodate various shapes of the human back. However, it is infeasible to balance the mechanism if only one cable is used to pull the discs. We designed a backbone using coiled steel tubings. The outer diameter is 2.5 mm and the inner diameter is 2 mm. It does not limit the degree of freedom due to its low stiffness. The backbone passes through all discs to address the balance problem. The force balance diagram of the discs is shown in Fig. 3, where (a) represents the case when the number of the discs is even and (b) represents the number of discs is odd. There is a slight difference in force balance between (a) and (b).

The assistive force of the back exoskeleton is transmitted from the cable to the human back. The cable pulls the distal disc, the $n^{th}$ disc, with the force, $F_c$. The $(n-1)^{th}$ disc and the backbone will generate the reaction forces to the $n^{th}$ disc, denoted by $F_{rn}$ and $F_{an}$. The discs can slide along the backbone, ensuring that the force $F_{an}$ passes through the center of the disc.

The condition of equilibrium for the three forces is that the directions of the forces pass through a single point and forces lie in the sagittal plane.

For the $n^{th}$ disc, all the forces $F_c$, $F_{rn}$ and $F_{an}$ will pass through the point $P_n$ and has the relationship:

$$F_{an} = F_c \tan\alpha, \ F_{rn} = F_c \sec\alpha, \ \alpha = \arctan(r_1/r_2) \quad (4)$$

$r_1$ and $r_2$ are the moment arms of $F_c$ and $F_{an}$ about the center of the spherical joint on the $n^{th}$ disc.

The other discs have similar force balance conditions like the distal disc. Taking the $(n-1)^{th}$ disc as an example, it is subjected to three forces, i.e. the reaction forces of the backbone, $F_{an-1}$, the $n^{th}$ disc, $F'_{rn}$, and the $(n-2)^{th}$ disc, $F_{rn-1}$. All of them pass through a single point, $P_{n-1}$, and have the relation:

$$F_{rn-1} = F'_{rn} = F_{rn}, F_{an-1} = 2F_{rn-1}\sin\alpha = 2F_{an} \quad (5)$$

In summary, for all the discs, the force balance conditions are

$$F_{an} = F_c \frac{r_1}{r_2}, F_{an-1} = F_{an-2} = \cdots = F_{a1} = 2F_c \frac{r_1}{r_2}$$
$$F_{rn} = F_{rn-1} = \cdots = F_{r1} = F_c \frac{\sqrt{r_1^2 + r_2^2}}{r_2} \quad (6)$$

The difference between the continuum mechanism with an odd and even number of discs lies in the different force balance conditions of the base. In the case of an odd number, the backbone will apply a reaction force, $F_{a0}$, to balance the base:

$$F_{a0} = F'_{r1} = F_{r1} = F_c \frac{\sqrt{r_1^2 + r_2^2}}{r_2} \quad (7)$$

While in the case of even case, the backbone will generate a reaction moment, $M$, besides the reaction force to render the base balanced:

$$F_{a0} = F'_{r1} = F_{r1} = F_c \frac{\sqrt{r_1^2 + r_2^2}}{r_2}, M = 2F_{a0}r_2 \quad (8)$$

From the above analysis, we see that for the under-actuated continuum mechanism, using one cable actuation and one backbone is efficient to balance the system in the sagittal plane. Moreover, compared to the traditional continuum mechanism [31], the proposed mechanism has the advantage that the backbone has no risk of instability because it is not subjected to end compression. In our design, the compression force along the human back is balanced by all the discs and transmitted to the base, which is located below the L5/S1 joint.

## IV. BIOMECHANICS MODELING OF HUMAN-ROBOT INTERACTION

With the kinematics and kinetic characterization of the robot mechanism, it is crucial to study biomechanics model of human-robot interaction to facilitate the development of assistive control of the soft exoskeleton.

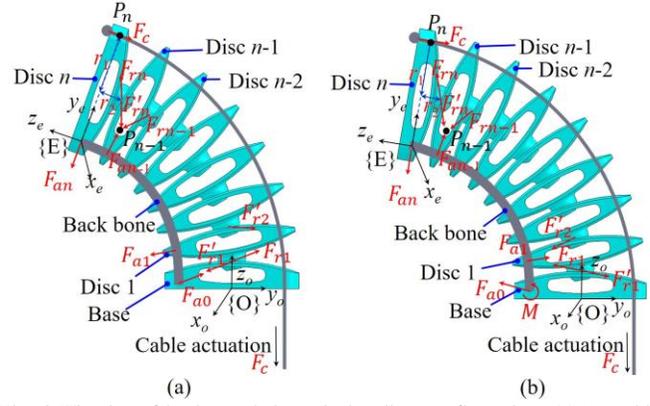

Fig. 3 Kinetics of back exoskeleton in bending configuration. (a) An odd number of discs. (b) An even number of discs. The exoskeleton assists human by one cable actuation and one backbone balancing the discs.

### A. Analytic Modeling of Human-Robot Interaction

The kinetic requirement of the back exoskeleton is to reduce the compression force and the shear force between discs which are the main causes of low back pain. Here we build a simple analytical model of the human spine to predict the effectiveness of the exoskeleton assistive force on reducing the forces in the human spine and muscle.

The bending model of the lumbar spine is simplified as one part that extends and flexes at the lumbar-sacral joint (L5/S1) in the sagittal plane. When the human lifts the load during stoop lifting described in Fig. 4, we can establish the static model in the flexed forward position to describe the relationship between the exoskeleton force and the forces in the human spine:

$$F_e D_e = -F_{exo} D_{exo} + m_{load} g D_{load} + m_{body} g D_{body} \quad (9)$$
$$F_p = F_e + m_{body} g \cos\theta + m_{load} g \cos\theta \quad (10)$$
$$F_s = -F_{exo} + m_{body} g \sin\theta + m_{load} g \sin\theta \quad (11)$$

$F_p$, $F_s$ denote the compressive force and shear force of intervertebral discs. $F_{exo}$ denotes the force applied by the back exoskeleton. $F_e$ denotes the muscle force of the lumbar. $m_{body}$ and $m_{load}$ are the mass of the human upper body and the load respectively. $D_{exo}, D_e, D_{load}, D_{body}$ are the moment arms of the exoskeleton, erector spinae muscle, load, and upper body respectively.

According to (9-11), it can be observed that if we increase the exoskeleton force, $F_{exo}$, the erector muscle force, $F_e$, the compressive force, $F_p$, and the shear force, $F_s$, decrease simultaneously. The gravity of the human and load are balanced partly by the assistive force of the exoskeleton. The static modeling is helpful for qualitative analysis of the effect of the exoskeleton on the human.

### B. Numerical Musculoskeletal Simulation of Human-Robot Interaction

We study numerical musculoskeletal modeling to characterize a more comprehensive human-robot physical interaction and simplify the conventional iterative exoskeleton design processes that heavily rely on prototype testing to inform design optimization. Moreover, the numerical model is convenient for rapid data-driven simulation using motion capture data. This has the potential to individualize the robot design and control strategy development. The numerical simulation of human-robot interaction aims to: 1) validate the feasibility of reducing the L5/S1 compression and shear force on a generic model of the human body; 2) evaluate the effect of our back exoskeleton.

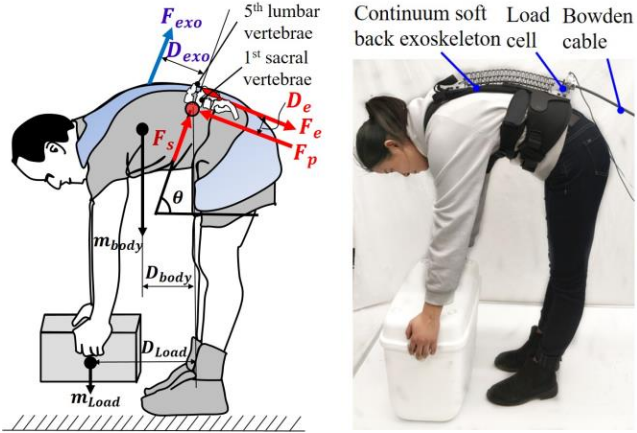

Fig. 4 Biomechanics model of human lifting. When the exoskeleton applies a force perpendicular to the human back, the spine compression force and shear force, and the muscle force decreases accordingly.

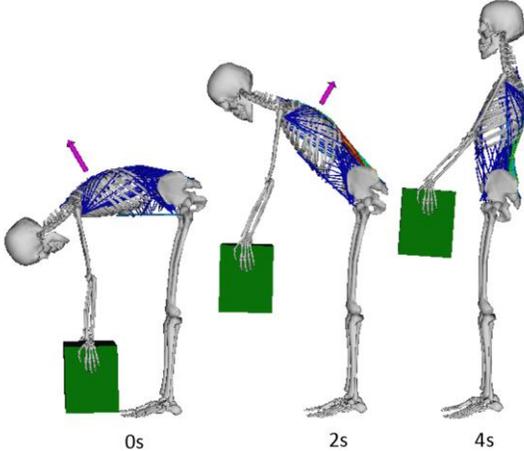

Fig. 5 Stoop lifting of the human body model. The arrow is the force applied at a different time (0,2,4 second) and the muscle color indicates its activation.

A highly detailed lumbar musculoskeletal model developed by Christopher et al. [32] was integrated with a generic full body model [33] and used for this study. The human body model as shown in Fig. 5 includes 34 body segments in the trunk, head, arms, and legs. The trunk region consisted of the head, cervical, thoracic and lumbar spines (5 lumbar vertebrae, L1 to L5), sacrum, and pelvis. The simulation was conducted in in-house musculoskeletal simulation software, CoBi-Dyn. During the simulation, the exoskeleton assistance force is at its maximum of 250N at the beginning (0 seconds) and gradually decreases to 0 at 4 seconds (at the erected position) following a cosine profile. As shown in Fig. 6, our simulation model predicts that the maximum disc compression force can be reduced by 37% (from 3751N to 2362N) during stoop lifting using the exoskeleton. Similarly, the maximum disc shear force can be reduced by 40% (from 561N to 336N) and the maximum average erector spinae muscle force can be reduced by 30% (from 33.5N to 23.5N) using our spine exoskeleton when 250N is used to pull the human back in fully flexed stoop position. This model serves as a design optimization tool that gives the designer a quantitative solution and flexibility to gauge the tradeoff between assistive torque and robot mass for human-robot interaction design.

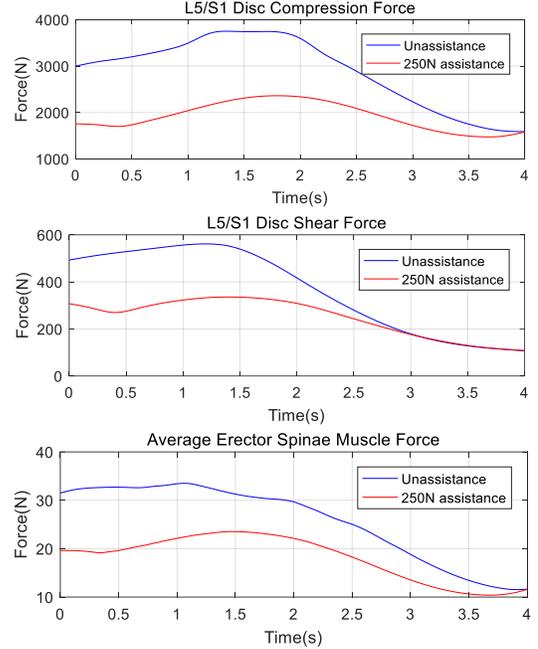

Fig. 6 Comparison L5-S1 disc compression and shear force in the situation with and without the back exoskeleton for 250N assistance. Our simulation model predicts that the maximum disc compression force can be reduced by 37% (from 3751N to 2362N), the maximum disc shear force can be reduced by 40% (from 561N to 336N) and the maximum average erector spinae muscle force can be reduced by30% (33.5N to 23.5N).

## V. ASSISTIVE CONTROL STRATEGIES

### A. Virtual Impedance Control for Back Assistance

We generate the assistive force using a virtual impedance model shown in Fig. 7. The assistive torque $T_r$ is generated by equation (12) from the desired position reference trajectory and the actual position trajectory. The desired trunk angle $\theta_r$, desired trunk angular velocity $\dot{\theta}_r$, and desired trunk angular acceleration $\ddot{\theta}_r$ are generated from a predefined desired position trajectory. We set the desired trajectory as 0 to create a virtual spring and damper that are connected to the ground. The trunk angle $\theta_a$, trunk angular velocity $\dot{\theta}_a$, and trunk angular acceleration $\ddot{\theta}_a$ were measured by an inertial measurement unit (IMU) sensor mounted on the trunk and the coordination was the same as the angle $\theta$ in Fig. 4. In our back exoskeleton, the cable force was controlled and applied to generate the assistive torque. Therefore, the assistive force we generated with the exoskeleton is given by equation (13).

$$T_r = J_d(\ddot{\theta}_a - \ddot{\theta}_r) + B_d(\dot{\theta}_a - \dot{\theta}_r) + K_d(\theta_a - \theta_r) \qquad (12)$$
$$F_r = T_r/r_1 \qquad (13)$$

### B. Control Architecture

The control architecture consists of four parts including a high-level controller, low-level controller, human-exoskeleton system, and wearable sensors shown in Fig. 8. 1) In the high-level controller, the virtual impedance mode is used to generate the force reference $F_r$ by the measured trunk angle $\theta_a$ input and the force control is a PID controller to track the force reference $F_r$ by the force error between the force reference $F_r$ and the measured cable force $F_r$. The control frequency in the high-level controller operated at 1000 Hz and is implemented in Matlab/Simulink Real Time. 2) In the low-level controller, a DSP microcontroller (TMS320F28335, Texas Instruments, USA) is used for the motor current and velocity control. It receives the velocity reference $V_r$ command by the CAN bus. The CAN bus based communication card (CAN-AC2-PCI, Softing Industrial Automation GmbH, USA) sends control command and acquires actuator state data from the low-level motor controller. The velocity controller implements a PID algorithm to track the angular velocity reference $\omega_r$ (by angular velocity error between the angular velocity reference $\omega_r$ and the measured motor angular velocity $\omega$). The current control also implements a PID algorithm to track the current reference $I_r$ (by the error between the current reference $I_r$ and the measured current $I$). 3) The human-exoskeleton model consists of motor, back exoskeleton, and human. The nominal torque of the motor is 2 Nm. The torque generated by the electric motor is transmitted to the cable by a 36:1 gear. The cable force pulls the back exoskeleton and produces assistive torque on the human. 4) The torque sensor is attached to the Bowden cable and is used to measure the interaction force between the back exoskeleton and the Bowden cable sheath, ($F_a$ in Fig. 8). The data acquisition (I/O) card (ADC, PCIe-6259, National Instrument, Inc., USA) acquires the load cell signals. An IMU is mounted on the subject trunk to measure the trunk motion (angle, angular velocity, angular acceleration) and the IMU data is transmitted to a target computer by serial port (RS-232).

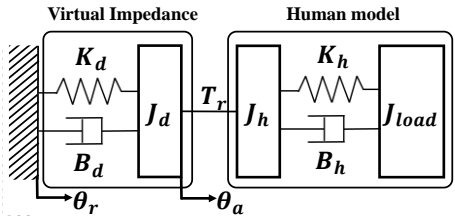

Fig. 7 Virtual impedance model. The assistive torque was generated by equation (12) from the desired position reference trajectory and the actual position trajectory with desired stiffness ($K_d$), damping ($B_d$), and inertia ($J_d$). Using the virtual impedance model, the exoskeleton generated an assistive torque reference $T_r$.

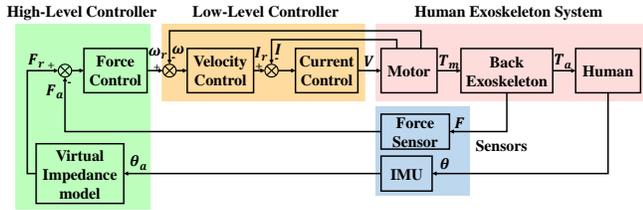

Fig. 8 The block diagram of back exoskeleton control for stoop assistance. It consisted of four parts: 1) the high-level controller generated the assistive force reference through the virtual impedance model and tracked the force reference 2) the low-level controller implemented the velocity and current control. 3) the human-exoskeleton system, and 4) the sensors measured the cable force and the human trunk motion.

## VI. EXPERIMENTS AND RESULTS

The experiment setup consists of the back exoskeleton, a tethered actuation platform, and a real-time control system, as shown in Fig. 9. The platform is equipped with a motor-gear-pulley transmission. The nominal speed of the motor is 1500 rpm, the gear ratio is 36:1, and the radius of the pulley is 0.05m. The platform can output a maximal 1500N pulling force and a 0.22 m/s translating speed for the cable. Currently, we used the tethered system to demonstrate proof of concept of our design and focus on control algorithm investigation by minimizing the impact of the mass of actuators and control electronics. The mass of the motor and gearbox are 274g and 290g respectively. Therefore, the actuator is lightweight to be potentially used in a portable version, which is now under development. Three subjects performed the stoop lifting of 15 kg with 10 repetitions. Each stoop cycle took 8 seconds that included 1) bending forward from stand up posture to trunk flexion for 4 seconds and 2) extending from trunk flexion to stand up posture for 4 seconds. The study was approved by the City University of New York Institutional Review Board, and all methods were carried out in accordance with the approved study protocol.

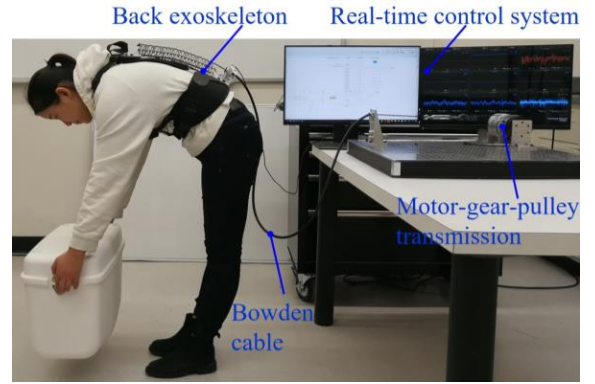

Fig. 9 A healthy subject wearing the exoskeleton performed stoop lifting with a 15 kg load. A tethered actuation platform provided cable actuation to power the continuum soft exoskeleton.

### A. Steerability Evaluation of the Continuum Exoskeleton

To test the relation between the cable displacement and the bending angles of the back exoskeleton, the cable is retracted from 5.23 cm to 0 cm causing the bending angle to change from 100° to 0°. The bending angle is defined as the angle between the end faces of the base and the top disc. The red lines in Fig. 10 are drawn to be parallel to the two end faces. The position of the center of the bending angle is the intersection of the two red lines. The steerability sequence of the back exoskeleton is depicted in Fig. 10. This demonstrates the feasibility of our robot to conform to human spine anatomy without limiting human movements.

### B. Stiffness Control of Back Exoskeleton

In this study, the desired stiffness is set as $200\sin(\theta_a)$ $(N)$, and the damping term is set as $20\dot{\theta}_a$ $(N)$, as in (14). The sine function is used because we intended to compensate the gravity term of the human and loading weight (which are related to $\sin(\theta_a)$).

$$F_r = 20\dot{\theta}_a + 200\sin(\theta_a) \qquad (14)$$

Fig. 11 illustrates the relationship between the motor current and the actual assistive force. It demonstrates the Bowden cable transmission system has hysteresis property but the force control is able to successfully compensate this nonlinear effect by feedforward control using the force measurement between the Bowden cable sheath and the exoskeleton. Fig. 12 depicts the relationship between the actual assistive force (blue line) and the desired spring assistive force. It demonstrates that the actual assistive force is highly consistent with the ideal spring assistive force and that the desired impedance model is achieved in our exoskeleton to assist stoop lifting.

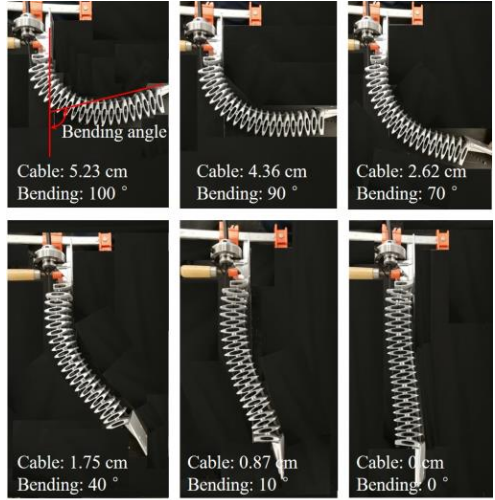

Fig. 10 Steerability demonstration of the continuum mechanism. When the cable is retracted from 5.23 cm to 0 cm, the bending angle is reduced from 100° to 0°.

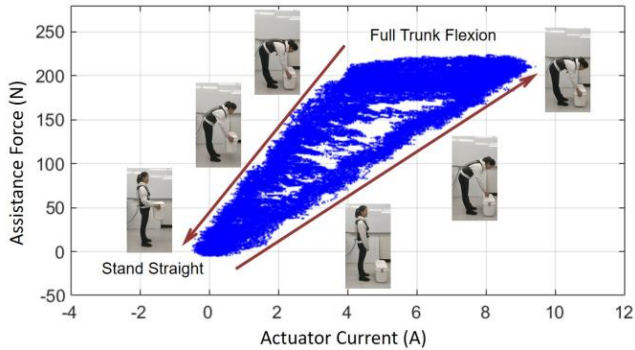

Fig. 11 The desired and actual assistive force during the stoop lifting. It demonstrates the hysteresis property due to the Bowden cable transmission mechanism. The hysteresis causes the open-loop assistive force control (that only implemented the current control) unsatisfactory tracking performance. In our control algorithm, we use feedforward control with a force sensor to directly measure the force between the exoskeleton and Bowden cable to achieve superior force tracking performance.

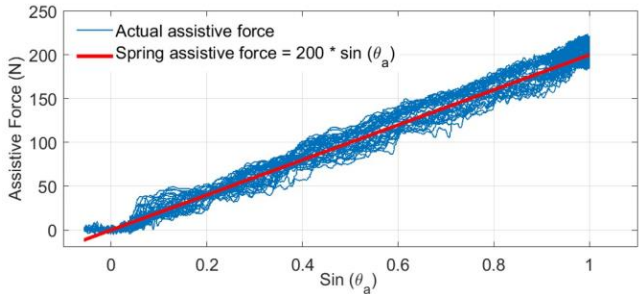

Fig. 12 The relationship between the assistive force and sine function of trunk angle $\sin(\theta_a)$ under stiffness control during stoop tasks in three subjects for total 30 stoop cycles. Compared to the desired spring assistive force and the actual assistive force, the two curves are highly linear and it demonstrates that the desired virtual impedance control can be performed well in our control system.

## C. Tracking Performance of Assistive Force Control

Fig. 13 illustrates the force control and the trunk angle variation during stoop tasks in three subjects for a total of 30 stoop cycles. The trunk angle was used to calculate the assistive torque by the virtual impedance model in equation (14). The mean of assistive force reference is annotated with a dashed blue line, the mean of actual assistive force is annotated with a red line, and the light blue area represents one standard deviation. The RMS error of force tracking is 6.63 N (3.3 % of the peak force 200N). Regardless of motion variability indicated by the standard deviation of trunk angles during 30 stoop cycles, our controller is able to successfully track the desired force with high accuracy.

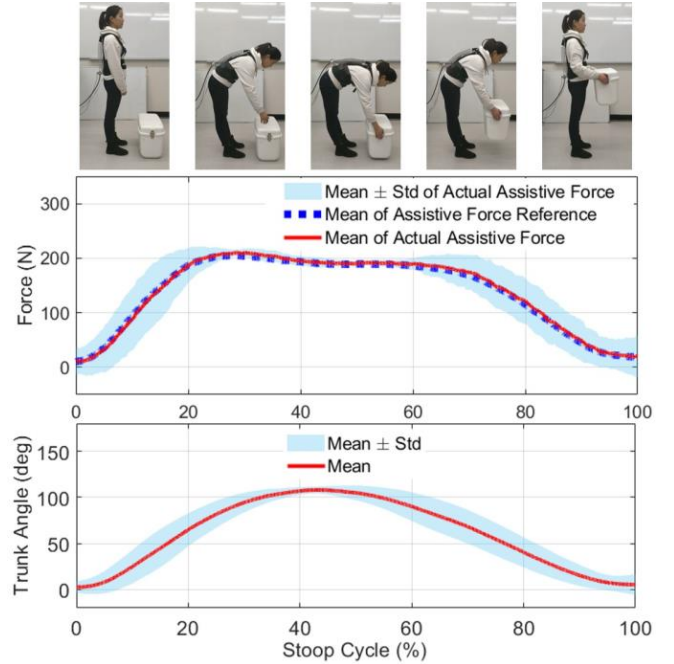

Fig. 13. Assistive force tracking performance and trunk angle measurement during stoop lifting. It was tested in three subjects and each subject performed 10 stoop cycles. The mean of actual assistive force (red line) tracked the mean of assistive force reference (blue dash line) well. The RMS error of force tracking in thirty stoop tasks was 6.63 N (3.3% of the peak force 200N).

## VII. CONCLUSION

Continuum soft exoskeletons represent a new design approach for wearable robots. It is particularly suitable for the assistance of articulations with either multiple segment structure (e.g. spine and fingers) or condyle joints (e.g. knee), or ball-and-socket joints (e.g. ankle and hip) as it helps avoid the misalignment between robotic joints and biological joints. The under-actuation nature of continuum soft exoskeleton ensures conformal adaptation of wearable robots to complex human anatomy. By studying the kinematics and kinetics modeling of the continuum soft exoskeleton, the design concept and the principle of assistance are revealed. We demonstrate that the back exoskeleton with one cable actuation can assist stoop lifting with less than 3.3% of tracking error while not restricting natural movement. In our further research, we will conduct a biomechanics study to quantify the benefit of exoskeleton-assisted lifting and compare it with the musculoskeletal simulation. We will further optimize the design of discs, backbone, and wearable structure to improve its ergonomics and enhance assistance performance.

ACKNOWLEDGMENT

This work is supported by the National Science Foundation grant NRI 1830613 and The City University of New York, City College. Any opinions, findings, and conclusions or recommendations expressed in this material are those of the author(s) and do not necessarily reflect the views of the funding organizations.